\documentclass[runningheads]{llncs}

\usepackage{eccv}

\usepackage{eccvabbrv}

\usepackage{graphicx}
\usepackage{booktabs}

\usepackage[accsupp]{axessibility}  

\usepackage{multirow}
\newcommand{\JS}[1]{{#1}}
\newcommand{\HJ}[1]{{#1}}
\newcommand{\AJ}[1]{{#1}}
\usepackage{wrapfig}
\usepackage{algorithm}
\usepackage{algpseudocode}
\usepackage{pifont}
\usepackage[table]{xcolor}
\usepackage{colortbl}

\usepackage{hyperref}

\usepackage{orcidlink}

\begin{document}

\title{Towards Sparsely Annotated Open-World Object Detection} 



\author{
HeeJu Han\orcidlink{0009-0005-4130-3387} \and
AJeong Kim\orcidlink{0009-0001-4946-6632} \and
Jinsun Park\thanks{Corresponding author}\orcidlink{0000-0002-2296-819X}
}

\authorrunning{H. Han et al.}



\institute{
Pusan National University, Republic of Korea
\\
\email{\{hhjjoa2817, jeonk636, jspark\}@pusan.ac.kr}
}

\maketitle


\begin{abstract}
  Real-world object detection operates under ambiguous supervision, where unlabeled regions may correspond to missing annotations of known objects or genuinely unknown categories.
  %
  \JS{
  These challenges have been addressed separately in Sparsely Annotated Object Detection (SAOD) and Open-World Object Detection (OWOD).
  }
  %
  %
  \HJ{
  In practice, their co-occurrence remains an open problem.
  }
  To address this problem, we introduce Sparsely Annotated Open-World Object Detection (SA-OWOD), a new task that jointly considers sparse supervision and the presence of unseen categories.
  We propose Dual-Perspective Object Discovery (DPOD), a unified framework that jointly models unlabeled known and unknown instances via two complementary mechanisms.
  The Known Target Recovery Module (KTRM) recovers supervision for unlabeled known instances and explicitly regularizes the feature space to separate known and unknown representations.
  Complementarily, the Dual-Disagreement Target Generator (DDTG) identifies reliable unknown candidates through cross-view semantic inconsistency.
  By integrating these modules, DPOD resolves contradictory supervision signals caused by ambiguous unlabeled regions.
  As a result, it prevents misclassification between known and unknown objects and stabilizes the decision boundaries.
  \JS{
  Experimental results on sparsely annotated open-world benchmarks demonstrate that the proposed method outperforms existing open-world detection methods, particularly in detecting unknown objects.
  }
  \HJ{The code is publicly available at: \url{https://github.com/HelloHeeju/SA-OWOD}}
  \keywords{Open-World Object Detection \and Sparsely Annotated Object Detection \and Ambiguous Supervision}
\end{abstract}


\section{Introduction}
\label{sec:intro}


Object detection inevitably operates under incomplete and semantically open conditions. 
In practice, many object instances remain unlabeled due to annotation costs and human limitations. 
However, most existing object detection methods~\cite{redmon2016you, girshick2014rich, carion2020end, howard2017mobilenets, cai2018cascade, lee2024crossformer} rely on two strong assumptions: (1) training datasets provide fully annotated labels without missing objects, and (2) the category set is closed, with no novel classes appearing after training. 
While recent progress has been driven by large-scale datasets~\cite{yu2020bdd100k, xia2018dota, kuznetsova2020open, cordts2016cityscapes, geiger2013vision} and advances in deep neural architectures, these assumptions rarely hold in realistic scenarios.


\begin{figure}[t]
    \centering
    \includegraphics[width=0.9\linewidth]{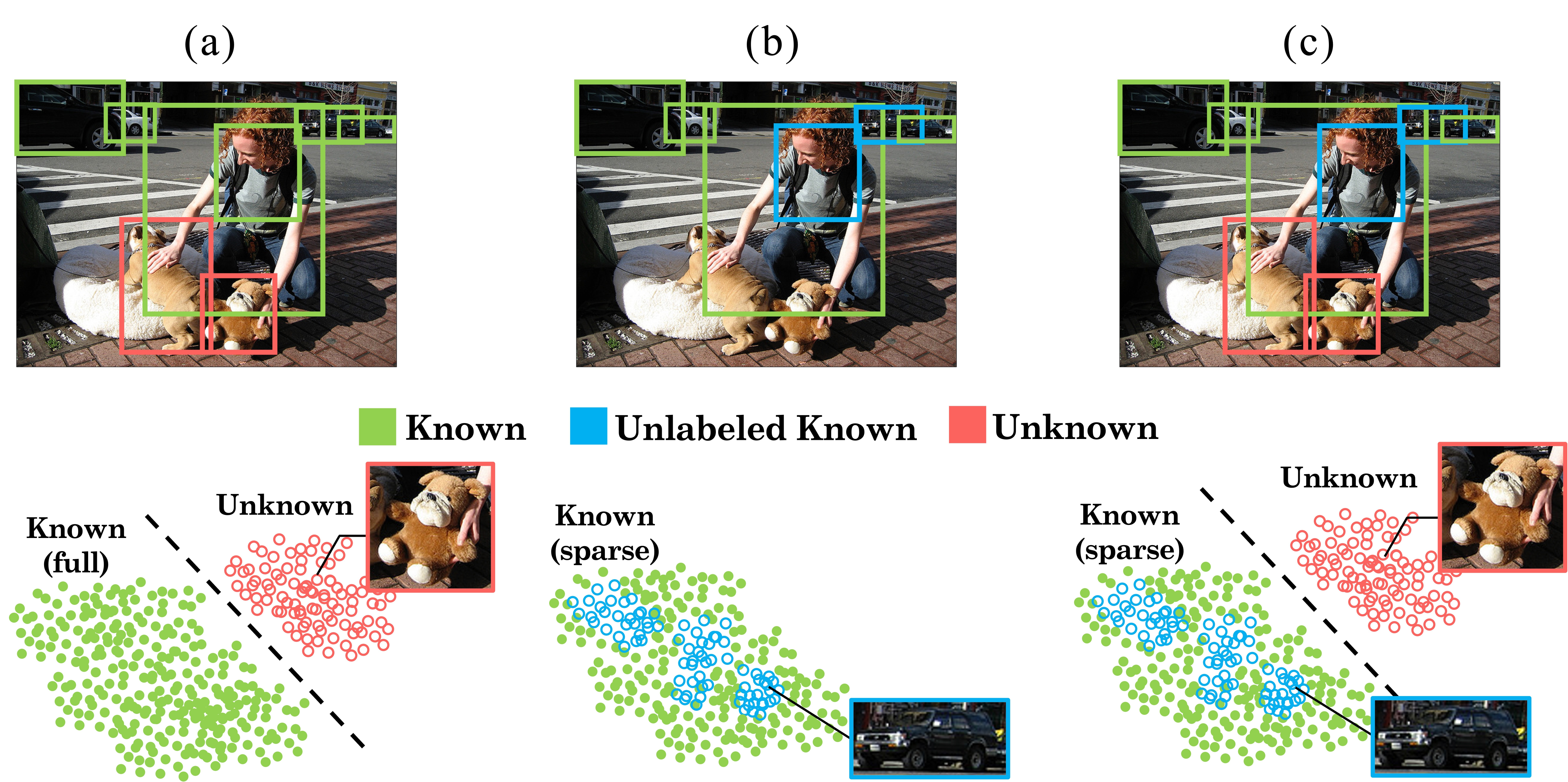}
    \caption{\textbf{Object detection paradigms for discovering unlabeled data.}
    (a) \textit{Open-World Object Detection} aims to identify novel objects as unknown, assuming full annotation data.
    (b) \textit{Sparsely Annotated Object Detection} focuses on recovering unlabeled known objects under sparse annotation settings, while assuming a closed set of categories.
    (c) \textit{Sparsely Annotated Open-World Object Detection} considers sparse annotations and open-world scenarios simultaneously, where unlabeled regions may correspond to either unlabeled known objects or unknown objects.
    }
    \label{fig:teaser}
\end{figure}


Open-World Object Detection (OWOD)~\cite{ma2023annealing, mullappilly2024semi, joseph2021ore, zohar2023prob, doan2024hyp} aims to not only detect known objects but also identify unseen ones as \textit{unknown}. 
To this end, existing OWOD methods adopt diverse strategies to model unknown objects, such as learning class-agnostic objectness scores and exploiting energy-based uncertainty measures. 
Nevertheless, these methods still depend on densely annotated training data, as illustrated in \cref{fig:teaser} (a).
%
\JS{
However, with sparse annotations where unlabeled known objects may exist, they are treated as background during training.
}
As a result, the detector may incorrectly suppress valid known instances or confuse them with unknown objects, which blurs the decision boundary between known and unknown classes and degrades the accuracy of both.

On the other hand, Sparsely Annotated Object Detection (SAOD)~\cite{wang2023calibrated, lee2025multispectral, yao2025pseudo, wu2018soft, lu2024progressive} addresses object detection under incomplete annotation settings, where only a subset of object instances \AJ{is} labeled per image during training (\ie, sparse annotations). 
In this setting, missing annotations cause detectors to incorrectly treat unlabeled objects as background regions, leading to false-negative supervision during training. 
To address this issue, several studies~\cite{wu2024co, wang2021co, zhou2022dense} attempt to compensate for annotation sparsity by generating pseudo-labels from unlabeled regions. 
However, these methods are fundamentally developed under a closed-world assumption, as shown in \cref{fig:teaser} (b). 
Under this assumption, every object belongs to a predefined set of known categories.
%
%
\HJ{
As a result, semantically novel objects outside the target classes are treated as background and therefore cannot be discovered.
}

Although SAOD and OWOD have been studied independently, real-world data contains both unlabeled known and unknown objects \AJ{simultaneously}. 
From the detector’s perspective, these instances appear as indistinguishable unlabeled regions during training. 
\HJ{
Without explicit supervision, the model cannot distinguish between unlabeled instances of known classes and unseen objects, leading to ambiguous training signals.
}
%
\JS{
To address this problem, we introduce a new task, \textit{Sparsely Annotated Open-World Object Detection (SA-OWOD)}, which assumes that unlabeled regions may contain both missing annotations of known categories and objects from unseen classes.
}
%
%
\JS{
As illustrated in \cref{fig:teaser} (c), SA-OWOD captures scenarios in which both types of unlabeled instances co-exist within a single image.
Unlike OWOD and SAOD, which address these challenges in isolation, SA-OWOD explicitly models their interplay.
}

Solving SA-OWOD requires a framework that can both recover unlabeled known objects and identify unknown objects. 
For this purpose, we introduce \textit{Dual-Perspective Object Discovery (DPOD)}. 
DPOD comprises two complementary components: the \textit{Known Target Recovery Module (KTRM)} and the \textit{Dual-Disagreement Target Generator (DDTG)}. 
\HJ{
KTRM recovers unlabeled known objects through pseudo-labeling and regularizes the decision boundary between known and unknown categories through feature-level alignment.
}
\HJ{
Complementing this, DDTG identifies additional unknown candidates by exploiting cross-view semantic disagreement.
}
Proposals with high objectness but low logit similarity are treated as reliable unknowns. 
By coupling pseudo-label recovery with disagreement-driven unknown discovery and feature-level regularization, DPOD resolves the inherent ambiguity of unlabeled regions in SA-OWOD.

The main contributions of our work are summarized as follows:
\begin{itemize}
    \item To the best of our knowledge, we are the first to formulate the Sparsely Annotated Open-World Object Detection (SA-OWOD) task. To address this challenge, we introduce a novel framework, DPOD.
    \item We propose two complementary modules, KTRM and DDTG, to resolve the ambiguity of unlabeled regions in SA-OWOD. KTRM recovers missing known objects to stabilize pseudo-label learning and refine decision boundaries, while DDTG discovers additional unknown instances via cross-view semantic disagreement.
    \item We construct sparsely annotated open-world benchmarks and an evaluation protocol that explicitly captures the dual nature of unlabeled regions. They allow quantitative evaluation of both missing known object recovery and unknown object discovery.
\end{itemize}


\section{Related Work}


\HJ{
Object detection has achieved remarkable progress, evolving from two-stage frameworks~\cite{ren2015faster, he2017mask} and efficient one-stage detectors~\cite{redmon2016you, lin2017focal} to recent transformer-based architectures~\cite{carion2020end, zhu2020deformable}. Despite their strong performance, these methods operate under a closed-world assumption — they recognize only predefined categories and treat all unlabeled or unknown objects as background. Consequently, these methods rely on static, fully annotated datasets. This limits their applicability to real-world scenarios, where annotations are often incomplete and novel categories continually emerge.
}


\subsection{Open-World Object Detection (OWOD)}


Open-World Object Detection (OWOD)~\cite{ma2023cat, wu2022two, wu2022uc, zhang2025open, yu2023open} aims to not only recognize known object categories but also identify novel (\textit{i.e.}, unknown) objects and incrementally learn them over time.

Pseudo-labeling methods identify unknown objects and generate pseudo-labels as additional supervision. 
ORE~\cite{joseph2021ore} introduces the OWOD setting and generates unknown pseudo-labels from high-objectness proposals. It further enhances unknown modeling using contrastive clustering and energy-based identification. 
OW-DETR~\cite{gupta2022ow} extends Deformable DETR~\cite{zhu2020deformable} with attention-guided pseudo-labeling and explicit supervision to distinguish unknown objects from background regions.

Class-agnostic methods focus on objectness rather than class-specific supervision, treating both known and unknown objects as foreground. 
PROB~\cite{zohar2023prob} introduces a probabilistic objectness head to generalize from known to unseen categories. 
RandBox~\cite{wang2023random} mitigates the bias of label-dependent proposal generation by adopting randomly generated region proposals. 
OrthogonalDet~\cite{sun2024exploring} decouples objectness and classification by enforcing orthogonality between objectness and class features. 
CROWD~\cite{majee2025looking} discovers diverse unknown instances via a submodular gain objective and then learns disentangled representations of known and unknown classes to reduce confusion.

However, despite their effectiveness, most existing OWOD methods still depend on densely annotated data. Under sparse annotation settings, unlabeled known objects obscure the boundary between known and unknown categories, leading to degraded detection performance.


\subsection{Sparsely Annotated Object Detection (SAOD)}


Sparsely Annotated Object Detection (SAOD)~\cite{han2018co, suri2023sparsedet, niitani2019sampling, yoon2021semi, zhang2020solving} addresses object detection when only a subset of object instances \AJ{is} annotated per \AJ{image}, resulting in sparse supervision. In such scenarios, detectors incorrectly interpret unlabeled objects as background regions, introducing false-negative supervision during training. 

To alleviate this issue, several studies leverage pseudo-labeling to compensate for missing annotations. uDenseTeacher~\cite{zhou2022dense} \AJ{utilizes} dense pseudo-labels generated from the teacher’s raw dense predictions instead of sparse box-level pseudo-labels with post-processing. 
Co-mining~\cite{wang2021co} jointly trains two detectors that exchange each other’s high-confidence predictions. 
Co-Student~\cite{wu2024co} adopts a collaborative strong–weak student framework. The two students leverage each other’s predictions, while a teacher model refines them to provide more reliable supervision.

However, these methods are designed under a closed-world assumption, where every object belongs to a predefined class set. As a result, objects from previously unseen categories are implicitly treated as background, preventing their detection.


\section{Method}


\subsection{SA-OWOD Problem Definition}



\begin{figure}[t]
    \centering
    \includegraphics[width=\linewidth]{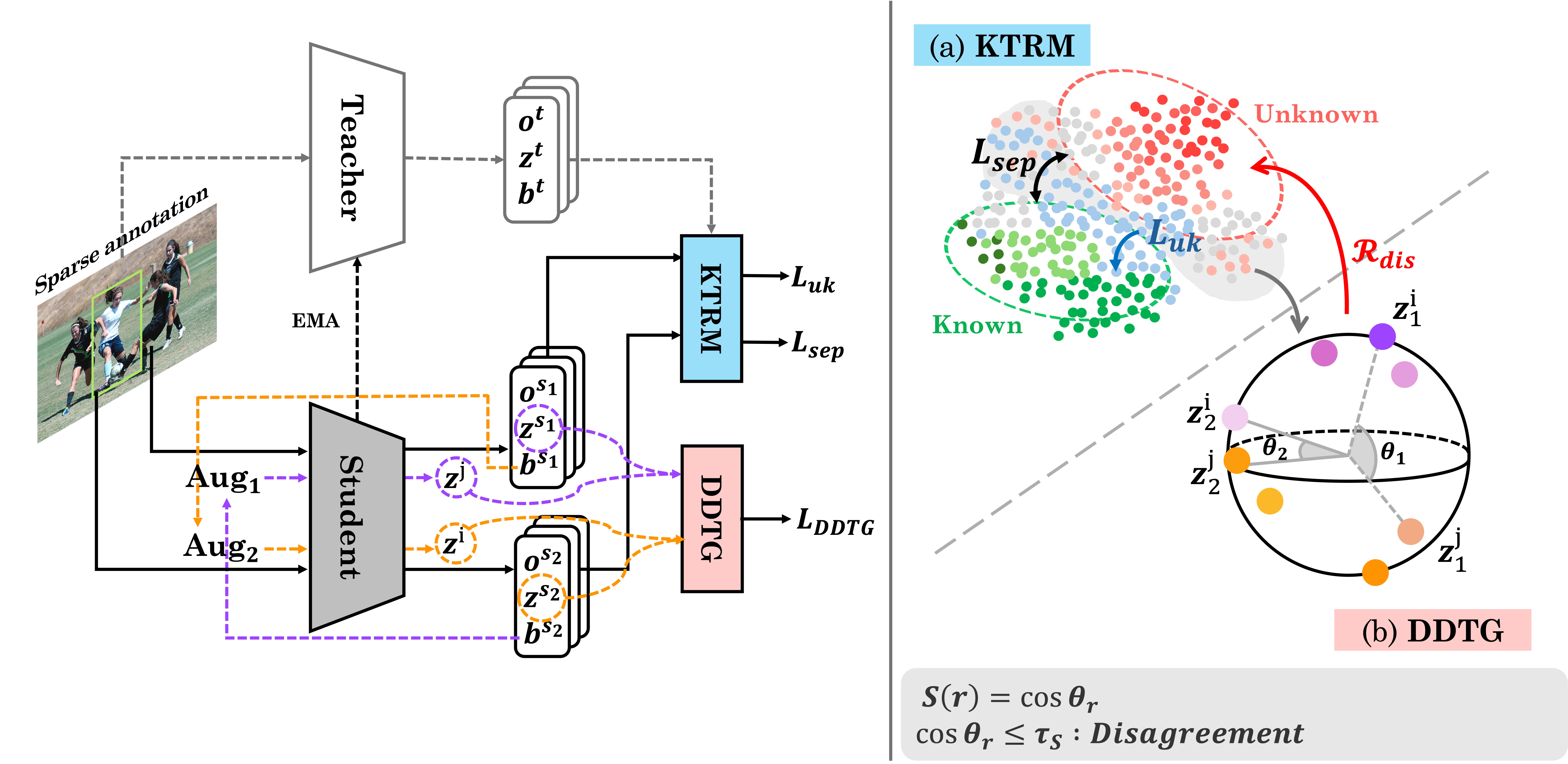}
    \caption{\textbf{Overview of DPOD.} (a) KTRM recovers supervision for unlabeled known objects and enforces known–unknown feature separation. (b) DDTG identifies additional unknown candidates via cross-view semantic inconsistency. The two signals jointly strengthen known–unknown separation and stabilize feature representations.}
    \label{fig:architecture}
\end{figure}

\JS{
In this work, we introduce a new task, SA-OWOD, which extends OWOD to a sparsely annotated scenario, as shown in \cref{fig:architecture}.
In the standard OWOD formulation~\cite{joseph2021ore}, a detector is trained to recognize known classes, identify unseen objects as \textit{unknown}, and incrementally incorporate them over time.
}
Unlike conventional OWOD, SA-OWOD assumes partial supervision within known classes: not all instances of known categories are annotated during training. Consequently, unlabeled known objects and truly unknown instances coexist without labels, yet must be assigned distinct semantic roles. This ambiguity constitutes the core challenge of SA-OWOD.


%
%
\JS{
Formally, the SA-OWOD is defined as follows:
At stage $t$, a detector $f_{t}$ is trained on a dataset $\mathcal{D}_{t} = \{\mathcal{I}_{t}, \mathcal{L}_{t}\}$, where $\mathcal{I}_{t}$ and $\mathcal{L}_{t}$ denote a set of training images and their corresponding partial annotations, respectively.
}
The annotations are provided only for the known class set $\mathcal{K}_t = \{1, 2, \dots, C\}$. However, not all instances belonging to $\mathcal{K}_t$ are annotated; a subset of known-class objects may remain unlabeled in $\mathcal{I}_t$. 
In addition, objects from the unknown class set $\mathcal{U} = \{C+1, C+2, \dots\}$ may appear without labels during training. Consequently, unlabeled regions may correspond to either missing annotations of known classes or objects from unknown categories.

Similar to OWOD, the problem follows an incremental learning protocol. At each stage, a subset of previously detected unknown classes $\bar{\mathcal{U}}_t \subset \mathcal{U}$ is annotated and incorporated into the known class set $\mathcal{K}_{t+1} = \mathcal{K}_t \cup \bar{\mathcal{U}}_t$. Due to memory and computational constraints, the model is updated from $f_t$ to $f_{t+1}$ using limited samples from previously known classes rather than retraining from scratch. This process is repeated over $T$ stages, requiring the detector to continuously discover novel objects while mitigating catastrophic forgetting.

In summary, SA-OWOD requires the detector to simultaneously: 
(1) recover supervision for unlabeled known instances and (2) detect truly unknown objects.


\subsection{Known Target Recovery Module (KTRM)}

\HJ{In SA-OWOD, unlabeled regions may correspond to known objects that were missed during annotation. 
If left unaddressed, these unlabeled known instances are treated as background during training. This introduces false-negative supervision and blurs the decision boundary between known and unknown categories. 
KTRM is designed to explicitly resolve this ambiguity through two complementary mechanisms:}
pseudo-label recovery for unlabeled known objects and representation-level separation between unlabeled known and unknown proposals. 
In sparsely annotated open-world settings, supervision for known classes is inherently incomplete, which makes classifier-based pseudo-labeling unreliable due to ambiguous decision boundaries. 
To alleviate this issue, we adopt a class-agnostic open-world detection framework based on CROWD~\cite{majee2025looking}, which produces object-centric proposals.

Based on these object-centric proposals, we recover supervision for unlabeled known objects using a pseudo-label generation strategy inspired by Co-Student~\cite{wu2024co}. 
A student detector first generates predictions from multiple augmented views of the same image. 
%
%
%
\JS{
These predictions are then refined by a teacher model through consistency-aware filtering, producing a reliable unlabeled known proposal set $\mathcal{R}_{uk}$.
We treat the elements of $\mathcal{R}_{uk}$ as pseudo-labels and combine them with the sparse ground-truth annotations to compensate for missing labels in known categories.
}
Each pseudo-label $p \in \mathcal{R}_{uk}$ provides a class label $y_{p}^{uk}$ and a bounding box $\mathbf{b}_{p}^{uk}$. 
%
%
\JS{
After that, each pseudo-label $p$ searches for the object proposal $r_{p}$ with the largest IoU and they are used for the training via classification loss $\ell_{cls}$ and regression loss $\ell_{reg}$.
}
The pseudo-label supervision is defined as:
\begin{equation}
    \mathcal{L}_{uk} =
    \frac{1}{|\mathcal{R}_{uk}|}
    \sum_{p \in \mathcal{R}_{uk}}
    \Big(
    \ell_{cls}\big( \mathbf{z}_{r_{p}}, y_{p}^{uk} \big)
    +
    \ell_{reg}\big( \mathbf{b}_{r_{p}}, \mathbf{b}_{p}^{uk} \big)
    \Big),
    \label{eq:uk_pseudo_label_loss}
\end{equation}
\JS{
where $\mathbf{z}_{r_{p}}$ and $\mathbf{b}_{r_{p}}$ denote class logits and bounding box of $r_{p}$, respectively.
}

While pseudo-label recovery compensates for missing supervision, feature representations of unlabeled known objects can still overlap with unknown instances in the embedding space. 
\HJ{
To enforce representation-level separation, we adopt a conditional gain objective inspired by the Facility-Location formulation in CROWD.
We adopt the Facility-Location formulation because it encourages each proposal to be separated from its nearest unknown counterpart rather than considering all unknown proposals simultaneously.
In contrast, distance-based objectives aggregate similarities across multiple unknown instances, which tends to blur local decision boundaries and weaken fine-grained discrimination.}

Let $\mathcal{R}_k$, $\mathcal{R}_{uk}$, and $\mathcal{R}_u$ denote labeled known 
\HJ{(from sparse ground-truth annotations)}, unlabeled known 
\HJ{(from pseudo-labeling)}, and unknown proposal sets, respectively. 
We define the unified known set as:
\begin{equation}
    \mathcal{A} = \mathcal{R}_k \cup \mathcal{R}_{uk}.
\label{eq:known_sets}
\end{equation}
Let $\mathcal{V}$ denote the set of all proposals.
The separation objective is formulated as a facility-location conditional gain:
\begin{equation}
\mathcal{L}_{sep} = - \frac{1}{|\mathcal{V}|} \sum_{i \in \mathcal{V}} \max \Big( 0, \max_{j \in \mathcal{A}} s_{ij} - \eta \, \max_{j \in \mathcal{R}_u} s_{ij} \Big),
\label{eq:sep_loss}
\end{equation}
\JS{
where $s_{ij}$ denotes a similarity metric between $i$-th and $j$-th proposals, and $\eta$ is a weighting factor that encourages separation between known and unknown features.
}

Minimizing $\mathcal{L}_{sep}$ encourages known proposals to move away from unknown regions in the embedding space, thereby reducing feature entanglement and stabilizing decision boundaries under sparse supervision.

The overall KTRM objective is defined as:
\begin{equation}
    \mathcal{L}_{KTRM} =
    \alpha_{uk} \, \mathcal{L}_{uk}
    +
    \alpha_{sep} \, \mathcal{L}_{sep},
    \label{eq:KTRM_loss}
\end{equation}
\JS{
where $\alpha_{uk}$ and $\alpha_{sep}$ are balancing hyperparameters.
}


\subsection{Dual-Disagreement Target Generator (DDTG)}


\begin{algorithm}[t]
\caption{Procedure for Selecting Unknown Candidates in DDTG}
\label{alg:ddtg}
\begin{algorithmic}[1]

\Require Detector $f_\theta$, 
two views $\{x^{s_1}, x^{s_2}\}$, 
objectness threshold $\tau_o$, similarity threshold $\tau_s$

\Ensure Disagreement proposal set $\mathcal{R}_{dis}$

\State Initialize $\mathcal{R}_{dis} \leftarrow \emptyset$

\For{each ordered pair $(x^i, x^j) \in \{(x^{s_1},x^{s_2}),(x^{s_2},x^{s_1})\}$}

    \State Obtain region proposals $\mathcal{R}^a$ from $x^a$

    \For{each proposal $r \in \mathcal{R}^a$}

        \State Project $r$ to $x^j$
        \State Extract logits 
        $\mathbf{z}_r^{i} = f_\theta(x^i, r)$,
        $\mathbf{z}_r^{j} = f_\theta(x^j, r)$
        
        \State Compute similarity
        $\mathcal{S}(r) =
        \frac{
        (\mathbf{z}_r^{i})^\top \mathbf{z}_r^{j}
        }{
        \|\mathbf{z}_r^{i}\|_2 \|\mathbf{z}_r^{j}\|_2
        }$
        
        \If{$o_r \ge \tau_o$ \textbf{and} $\mathcal{S}(r) \le \tau_s$}
            \State $\mathcal{R}_{dis} \leftarrow \mathcal{R}_{dis} \cup \{r\}$
        \EndIf

    \EndFor

\EndFor

\State \Return $\mathcal{R}_{dis}$

\end{algorithmic}
\end{algorithm}


In open-world settings, unknown objects often manifest as regions where semantic predictions are inherently unstable~\cite{sun2024exploring}. Unlike known categories, unknown objects tend to reside near ambiguous decision boundaries. Motivated by this observation, we introduce the Dual-Disagreement Target Generator (DDTG), summarized in Algorithm~\ref{alg:ddtg}. DDTG identifies unknown object candidates by measuring cross-view semantic inconsistency. Such inconsistency typically arises from unstable regions in the classification space associated with unknown objects.

Given two views of the same image, the detector produces class logits for each object proposal. We compare the class logit vectors obtained from the two views to identify prediction disagreement. However, reliable comparison requires that the corresponding \AJ{regions of interest} (RoIs) refer to the same spatial region. Naively matching proposals by intersection-over-union (IoU) can be unstable, as even minor localization shifts may lead to substantially different RoI features. To ensure consistent cross-view comparison, we generate each RoI from a reference view and project it to the other view to extract geometrically aligned RoI features. Prediction disagreement is then quantified by computing the cosine similarity between the aligned class logit vectors.

Given a proposal $r$, let $\mathbf{z}_r^i \in \mathbb{R}^C$ and $\mathbf{z}_r^j \in \mathbb{R}^C$ denote the classification logits predicted from two different views. We measure their semantic consistency using cosine similarity:
\begin{equation}
    \mathcal{S}(r) = 
    \frac{
    \left(\mathbf{z}_r^{i}\right)^{\top} \mathbf{z}_r^{j}
    }{
    \|\mathbf{z}_r^{i}\|_2 \, \|\mathbf{z}_r^{j}\|_2
    }.
\label{eq:cos_similarity}
\end{equation}
The set of disagreement proposals is defined as:
\begin{equation}
    \mathcal{R}_{dis} = \Big\{ r \in \mathcal{R} \;\Big|\; o_r \ge \tau_o, \;\mathcal{S}(r) \le \tau_s \Big\},
\label{eq:proposal_dis}
\end{equation}
%
%
\JS{
where \(o_r\) denotes the objectness score of proposal \(r\), $\tau_{o}$ is the objectness threshold, and $\tau_{s}$ is the similarity threshold.
}
Only proposals that are sufficiently object-like but exhibit low semantic consistency are treated as disagreement candidates. Similar to KTRM, we apply both classification and separation objectives to the disagreement set in order to enforce consistent unknown prediction. The resulting DDTG loss is defined as: 
\begin{equation}
\begin{split}
    \mathcal{L}_{DDTG} = \;&
    \beta_{cls} \,
    \frac{1}{|\mathcal{R}_{dis}|}
    \sum_{d \in \mathcal{R}_{dis}}
    \ell_{cls}\big( \mathbf{z}_{r_d}, y_d^{dis} \big) \\
    &-
    \beta_{sep} \,
    \frac{1}{|\mathcal{V}|}
    \sum_{i \in \mathcal{V}} \max \Big( 0, \max_{j \in \mathcal{A}} s_{ij} - \eta\, \max_{j \in \mathcal{R}_{dis}} s_{ij} \Big),
\label{eq:DDTG_loss}
\end{split}
\end{equation}
\JS{
where $y_{d}^{dis}$ denotes the unknown class label, and $\mathbf{z}_{r_d}$ denotes the class logits of the proposal $r_d$ that best matches the disagreement candidate $d$. $\beta_{cls}$ and $\beta_{sep}$ are balancing hyperparameters.
}

DDTG provides an additional supervisory signal for unknown object learning by modeling cross-view semantic instability. 
While objectness filtering ensures that candidate regions correspond to potential objects, it does not capture ambiguity in class predictions under view perturbations. By explicitly measuring disagreement between aligned logits, DDTG complements objectness-based supervision and enhances the robustness of unknown object discovery.

Accordingly, the overall loss for unlabeled object supervision is defined as the combination of the KTRM loss and the DDTG loss:
\begin{equation}
    \mathcal{L}_{DPOD} =
    \gamma_{KTRM} \, \mathcal{L}_{KTRM}
    +
    \gamma_{DDTG} \, \mathcal{L}_{DDTG},
\label{eq:DPOD_loss}
\end{equation}
\JS{
where $\gamma_{KTRM}$ and $\gamma_{DDTG}$ are balancing hyperparameters.
}


\section{Experiments}


\subsection{Datasets and Sparse Annotation Settings}
\label{sec:datasets}


\begin{table}[t]
    \centering
    \caption{\textbf{Task composition in open-world evaluation protocol.} The semantics of each task and the number of images and instances (objects) across splits are shown.}
    
    \setlength{\tabcolsep}{4pt}
    \renewcommand{\arraystretch}{1.15}
    
    \resizebox{0.8\columnwidth}{!}{%
    \begin{tabular}{p{2.8cm} | c c c c}
    \toprule
     & Task 1 & Task 2 & Task 3 & Task 4 \\
    \midrule
    Semantic split
    & \parbox[c]{1.6cm}{\centering VOC\\Classes}
    & \parbox[c]{2.8cm}{\centering Outdoor, Accessories,\\Appliance, Truck}
    & \parbox[c]{1.8cm}{\centering Sports,\\Food}
    & \parbox[c]{2.8cm}{\centering Electronic, Indoor,\\Kitchen, Furniture} \\
    \midrule
    \# training images  & 16,551 & 45,520 & 39,402 & 40,260 \\
    \# test images      & 4,952  & 1,914  & 1,642  & 1,738 \\
    \# train instances  & 47,223 & 113,741 & 114,452 & 138,996 \\
    \# test instances   & 14,976 & 4,966  & 4,826  & 6,039 \\
    \bottomrule
    \end{tabular}%
    }
    \label{tab:owod_task_split}
\end{table}


\begin{wrapfigure}{r}{0.4\linewidth}
    \centering
    \includegraphics[width=\linewidth]{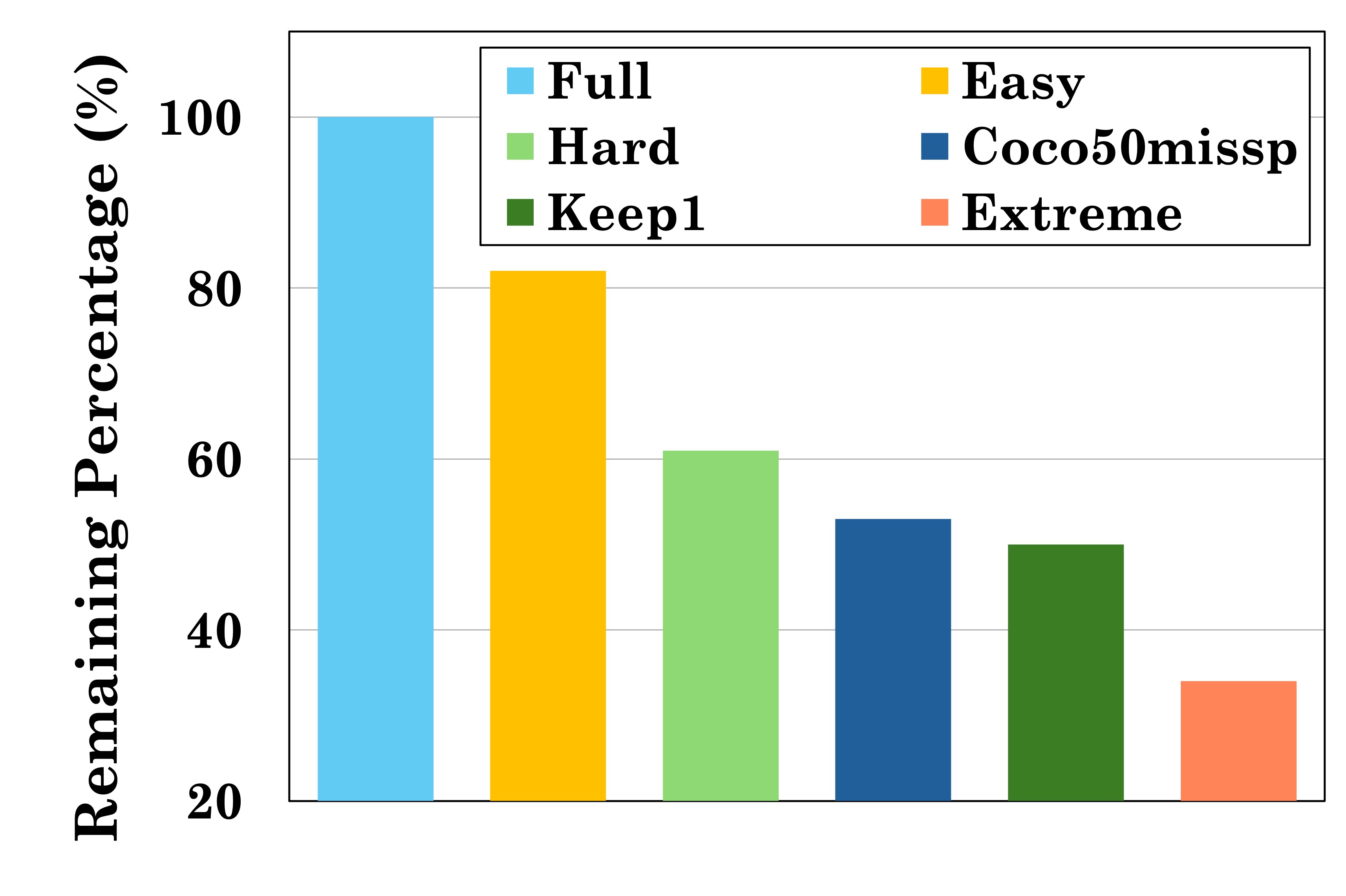}
    \caption{\textbf{Remaining annotation ratio under each sparse configuration (all tasks aggregated).}}
    \label{fig:dataset_ratio}
\end{wrapfigure}


We utilize the Open-World Object Detection dataset~\cite{joseph2021ore}, which is constructed from Pascal VOC~\cite{everingham2010pascal} and MS-COCO~\cite{lin2014microsoft}. Following the standard OWOD dataset protocol, we organize all VOC classes as the first task $T_1$, while the remaining 60 COCO classes are divided into three successive tasks. This division introduces semantic drifts between tasks (see \cref{tab:owod_task_split}). During the training of task $T_t$, classes from previous tasks $\{T_\tau : \tau < t\}$ are treated as \textit{known}, while classes from future tasks $\{T_\tau : \tau > t\}$ are treated as \textit{unknown}. This setup enables progressive learning under an open-world scenario, where the model incrementally encounters new classes while retaining knowledge of previously learned ones.

To simulate sparse annotation scenarios, we consider five training configurations~\cite{zhang2020solving, li2022siod} designed to reduce available annotations. Unlike conventional sparse annotation settings that remove annotations globally across the entire dataset, our framework follows an open-world protocol in which different tasks introduce different sets of classes. Accordingly, annotation removal is applied independently within each task. For example, under the Easy setting, one annotation is removed per image for each task individually: one annotation is removed for classes in Task 1, another for classes in Task 2, and so on. This ensures that each task observes its own sparse annotation setting, while images shared across tasks may have multiple annotations removed corresponding to the task-specific classes.

\JS{
\AJ{Specifically}, five configurations are defined based on the number of retained annotations per image per task:
}
\begin{itemize}
    \item \textbf{Easy}: Remove a single annotation, retaining at least one annotation per class in the current task; approximately 18.3\% of all labels are removed.
    \item \textbf{Hard}: Remove half of the annotations, retaining at least one annotation per class in the current task; approximately 38.8\% of all labels are removed.
    \item \textbf{Coco50missp}: Remove half of the annotations; approximately 46.7\% of labels are removed. Unlike the Hard setting, this does not guarantee at least one annotation per class in each image.    
    \item \textbf{Keep1}: Keep only one annotation for each category present in an image; approximately 50.0\% of labels are removed.
    \item \textbf{Extreme}: Retain only a single annotation; approximately 65.8\% of labels are removed.
\end{itemize}

For evaluation, we adopt the Pascal VOC test split and the MS COCO validation split.


\subsection{Implementation Details}


In this study, we adopt a Faster R-CNN~\cite{ren2015faster} as the baseline~\cite{majee2025looking}. We employ a ResNet50~\cite{he2016deep} backbone pre-trained on ImageNet~\cite{deng2009imagenet}. The network is optimized using AdamW~\cite{loshchilov2017decoupled} with a learning rate of $2.5\times10^{-5}$ and a weight decay of $1\times10^{-4}$. For each task, the model is trained for 15,000 iterations, followed by 15,000 iterations of incremental learning, resulting in a total of 30,000 iterations (approximately 9 epochs). All experiments are conducted on four NVIDIA RTX A6000 GPUs with a batch size of 12. All weighting coefficients $\alpha_{uk}$, $\alpha_{sep}$, \AJ{$\beta_{cls}$},
$\beta_{sep}$, $\gamma_{KTRM}$, and $\gamma_{DDTG}$ are set to 1, and the facility-location parameter $\eta$ is set to 1. During inference, predictions are obtained from 10,000 pre-defined bounding boxes, after which Non-Maximum Suppression (NMS)~\cite{neubeck2006efficient} is applied to remove redundant detections.
\HJ{Our model comprises 106.2M parameters with a computational cost of 965.4 GFLOPs, running at 2.52 FPS with a latency of 396.1 ms.}


\definecolor{ourscolor}{RGB}{230, 240, 255}

\begin{table*}[t!]
\centering
\caption{\textbf{Sparsely Annotated Open-World Object Detection (SA-OWOD) results.} Existing open-world detectors are benchmarked on sparse annotation datasets for fair comparison. K-mAP and U-Recall denote Known mAP and Unknown Recall, respectively.}

\begin{subtable}[t]{\textwidth}
    \centering
    \resizebox{\textwidth}{!}{\renewcommand{\arraystretch}{1.15}
    \setlength{\tabcolsep}{6pt}
    \begin{tabular}{c|l|cc|cc|cc|c}
    \toprule
    \multirow{2}{*}{Set} & \multirow{2}{*}{Method}
    & \multicolumn{2}{c|}{\textbf{Task 1}}
    & \multicolumn{2}{c|}{\textbf{Task 2}}
    & \multicolumn{2}{c|}{\textbf{Task 3}}
    & \multicolumn{1}{c}{\textbf{Task 4}} \\
    \cmidrule(lr){3-9}
    & & K-mAP & U-Recall
    & K-mAP & U-Recall
    & K-mAP & U-Recall
    & K-mAP \\
    \midrule
    \multirow{4}{*}{Full} 
        & RandBox~\cite{wang2023random} & 61.8 & 10.6 & 45.3 & 6.3 & 39.4 & 7.8 & 35.4 \\
        & PROB~\cite{zohar2023prob} & 59.5 & 19.4 & 44.0 & 17.4 & 36.0 & 19.6 & 31.5 \\
        & OrthogonalDet~\cite{sun2024exploring} & 61.3 & 24.6 & 47.0 & 26.3 & 41.3 & 29.1 & 37.9 \\
        & CROWD~\cite{majee2025looking} & 61.7 & 57.9 & 47.8 & 53.6 & 42.5 & 69.6 & 38.5 \\
    \midrule
    \multirow{5}{*}{Easy} 
        & RandBox & 49.91 & 4.90 & 38.36 & 4.30 & 33.12 & 4.60 & 29.79 \\
        & PROB & 52.83 & 18.27 & 38.35 & 15.35 & 32.64 & 19.1 & 29.06 \\
        & OrthogonalDet & 56.19 & 17.05 & \textbf{41.91} & 22.51 & 36.73 & 25.13 & \textbf{34.60} \\
        & CROWD & 53.28 & 45.53 & 38.24 & 30.22 & 35.43 & 53.34 & 32.28 \\
        & \cellcolor{ourscolor}\textbf{Ours} & \cellcolor{ourscolor}\textbf{56.48} & \cellcolor{ourscolor}\textbf{55.34} & \cellcolor{ourscolor}39.53 & \cellcolor{ourscolor}\textbf{52.47} & \cellcolor{ourscolor}\textbf{36.90} & \cellcolor{ourscolor}\textbf{63.68} & \cellcolor{ourscolor}32.15 \\
    \midrule
    \multirow{5}{*}{Hard} 
        & RandBox & 49.96 & 2.69 & 27.66 & 3.62 & 29.86 & 3.72 & 27.26 \\
        & PROB & 50.18 & 18.26 & 36.24 & 13.91 & 30.98 & 17.57 & 27.01 \\
        & OrthogonalDet & 53.14 & 19.19 & \textbf{38.03} & 18.28 & 32.87 & 24.44 & \textbf{32.14} \\
        & CROWD & 49.29 & 49.80 & 34.83 & 33.35 & 33.63 & 50.80 & 29.91 \\
        & \cellcolor{ourscolor}\textbf{Ours} & \cellcolor{ourscolor}\textbf{54.09} & \cellcolor{ourscolor}\textbf{51.85} & \cellcolor{ourscolor}33.36 & \cellcolor{ourscolor} \textbf{48.25} & \cellcolor{ourscolor}\textbf{35.79} & \cellcolor{ourscolor}\textbf{64.00} & \cellcolor{ourscolor}31.15 \\
    \midrule
    \multirow{5}{*}{Coco50missp} 
        & RandBox & 45.43 & 0.63 & 28.02 & 0.76 & 16.93 & 1.47 & 21.09 \\
        & PROB & 48.84 & 18.27 & 34.58 & 14.46 & 28.96 & 16.96 & 25.65 \\
        & OrthogonalDet & \textbf{54.38} & 15.41 & \textbf{35.56} & 25.16 & \textbf{31.26} & 29.24 & \textbf{28.56} \\
        & CROWD & 49.90 & 43.91 & 32.03 & 24.14 & 29.92 & 41.12 & 26.13 \\
        & \cellcolor{ourscolor}\textbf{Ours} & \cellcolor{ourscolor}53.78 & \cellcolor{ourscolor}\textbf{51.39} & \cellcolor{ourscolor}30.73 & \cellcolor{ourscolor}\textbf{47.69} & \cellcolor{ourscolor}30.90 & \cellcolor{ourscolor}\textbf{59.00} & \cellcolor{ourscolor}24.74 \\
    \midrule
    \multirow{5}{*}{Keep1} 
        & RandBox & 44.35 & 2.95 & 31.34 & 3.70 & 28.72 & 4.18 & 25.05 \\
        & PROB & 48.81 & 17.84 & 34.67 & 14.06 & 29.59 & 16.95 & 26.47 \\
        & OrthogonalDet & \textbf{50.47} & 14.07 & \textbf{35.00} & 16.60 & 31.58 & 19.71 & \textbf{30.28} \\
        & CROWD & 46.45 & 44.59 & 32.76 & 33.27 & 32.16 & 52.07 & 28.34 \\
        & \cellcolor{ourscolor}\textbf{Ours} & \cellcolor{ourscolor}47.97 & \cellcolor{ourscolor}\textbf{48.03} & \cellcolor{ourscolor}33.09 & \cellcolor{ourscolor}\textbf{48.28} & \cellcolor{ourscolor}\textbf{32.37} & \cellcolor{ourscolor}\textbf{52.22} & \cellcolor{ourscolor}28.21 \\
    \midrule
    \multirow{5}{*}{Extreme} 
        & RandBox & 36.62 & 0.68 & 21.93 & 1.50 & 19.55 & 2.39 & 17.95 \\
        & PROB & 37.78 & 17.86 & 26.09 & 14.83 & 23.41 & 17.60 & 19.38 \\
        & OrthogonalDet & 41.39 & 12.41 & \textbf{26.20} & 14.77 & 23.30 & 24.62 & \textbf{22.31} \\
        & CROWD & 34.37 & 40.66 & 24.84 & 31.17 & 24.46 & 48.64 & 20.47 \\
        & \cellcolor{ourscolor}\textbf{Ours} & \cellcolor{ourscolor}\textbf{45.37} & \cellcolor{ourscolor}\textbf{45.56} & \cellcolor{ourscolor}25.98 & \cellcolor{ourscolor}\textbf{44.32} & \cellcolor{ourscolor}\textbf{26.58} & \cellcolor{ourscolor}\textbf{57.53} & \cellcolor{ourscolor}20.19 \\
    \bottomrule
    \end{tabular}
    }
\end{subtable}

\label{tab:ow_main_result}
\end{table*}


\subsection{Main Results}


\textbf{SA-OWOD Performance.}
Table~\ref{tab:ow_main_result} compares existing OWOD methods under sparse annotations. 
All methods are retrained under identical sparse annotation protocols for fair comparison. 
K-mAP denotes the mean Average Precision over known categories, and U-Recall evaluates unknown object discovery. 
As annotation sparsity increases, prior OWOD approaches exhibit substantial degradation, particularly in U-Recall. 
In contrast, our method achieves superior U-Recall while maintaining competitive K-mAP across all sparsity levels.

\begin{figure}[t]
    \centering
    \includegraphics[width=\linewidth]{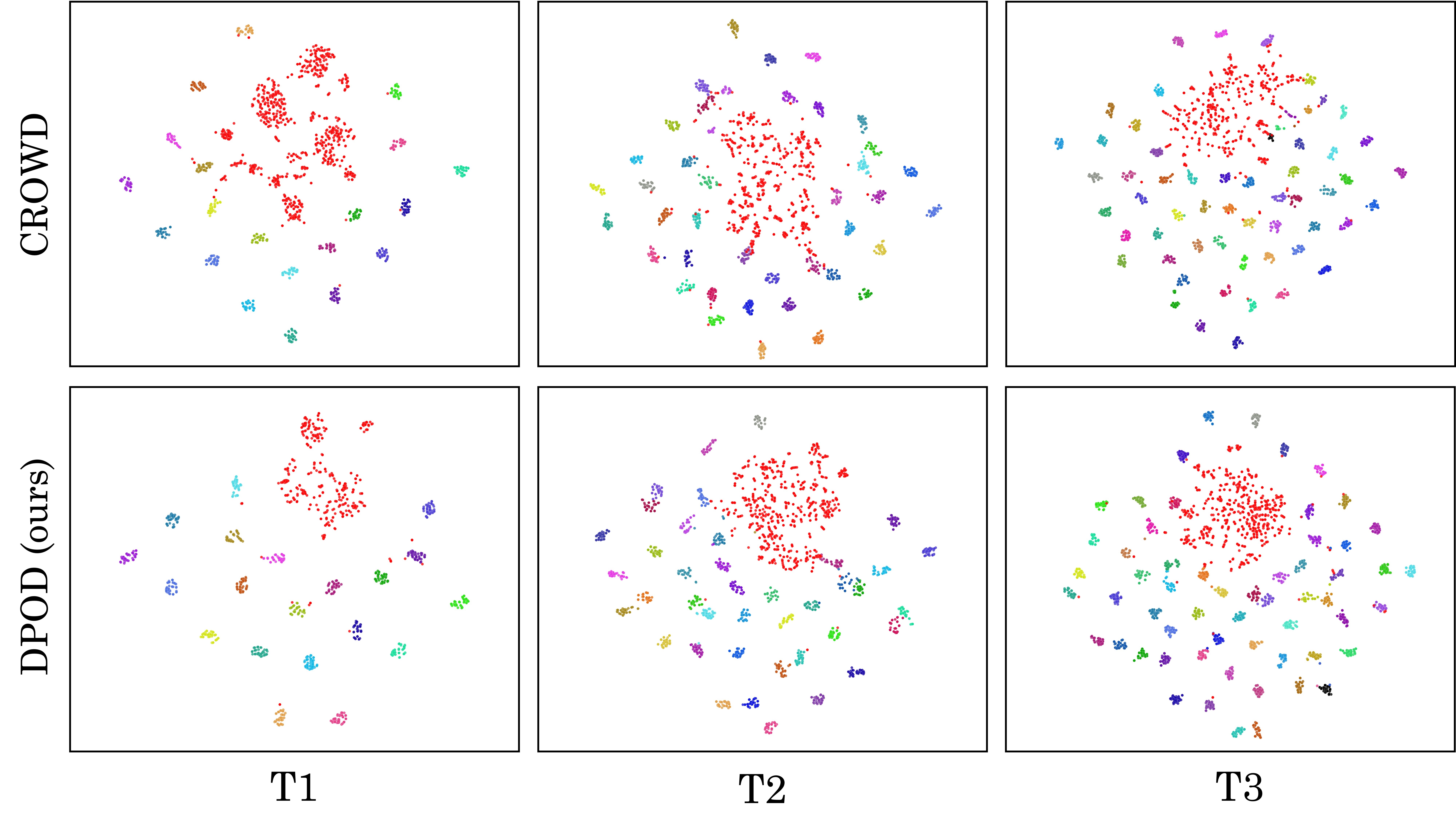}
    \caption{\textbf{t-SNE visualization of feature embeddings.}
        The distributions of known and unknown samples are shown for Tasks 1--3.
        Known classes are represented by different colors, while unknown objects are shown in \textcolor{red}{red}.
        Our method shows clearer separation between known and unknown samples compared to the baseline.}
    \label{fig:tsne}
\end{figure}

%
\JS{
In the Easy setting, our method improves U-Recall by 9.81\%p over CROWD and by 38.29\%p over OrthogonalDet in Task 1, achieving the best K-mAP in Tasks 1 and 3. 
}
These results demonstrate that our framework improves unknown discovery without sacrificing known-class detection.
As illustrated in \cref{fig:tsne}, the baseline tends to mix unknown samples with known clusters in the feature space. 
This phenomenon arises because sparsely annotated datasets often contain unlabeled known objects that are unintentionally treated as unknown during training, leading to ambiguous feature representations.
KTRM recovers these via objectness-guided pseudo-labeling, preventing them from being incorrectly treated as unknown samples.
Simultaneously, DDTG leverages cross-view semantic disagreement to identify reliable unknown candidates. 
By disentangling unlabeled known and unknown instances, our framework reduces the boundary confusion that often degrades prior OWOD methods under sparse supervision.

As sparsity intensifies (\ie, Easy $\rightarrow$ Hard $\rightarrow$ Extreme), the performance gap widens further.
In the Hard setting, our method surpasses CROWD by 14.90\%p and OrthogonalDet by 29.97\%p in Task 2. 
Under the Extreme sparsity, the unknown recall gap over OrthogonalDet exceeds 29.55\%p in multiple tasks. 
This robustness indicates that KTRM effectively compensates for missing supervision, while DDTG prevents premature consolidation of ambiguous proposals into known categories.


\noindent\textbf{Challenging Stages.}
In Task 2, our K-mAP is slightly lower than OrthogonalDet due to integrating novel classes under incomplete supervision, which causes unstable boundaries. 
\HJ{
However, DDTG improves robustness to ambiguity by delaying overconfident assignments under weak cross-view semantic consistency. This results in a 29.97\%p gain in U-Recall over OrthogonalDet in Hard Task 2.
} 
%
\JS{
In Extreme Task 2, OrthogonalDet achieves the highest absolute K-mAP\AJ{;} however, its U-Recall drops by 43.84\% compared to the fully annotated setting.
}
In contrast, although our method exhibits a comparable K-mAP reduction, the degradation in U-Recall is limited to only 17.31\%. 
\HJ{This gap demonstrates that our framework achieves a better balance between known detection and unknown discovery under sparse annotations.}

In contrast, Task 4, a later incremental stage, has few remaining unknowns.
Consequently, the amount of supervision related to unknown discovery becomes limited, and the learning objective shifts toward fine-grained discrimination among many known categories. 
Since our framework is primarily designed to resolve ambiguity between unlabeled known and unknown objects, the relative contribution of disagreement-driven unknown modeling (DDTG) naturally decreases when unknown signals become scarce. 
Therefore, the smaller performance margin observed in Task 4 reflects a change in task characteristics rather than a limitation in modeling class separation. 
Even under this shift, our method maintains competitive performance, demonstrating stable behavior across incremental learning stages.

\noindent\textbf{Discussion.}
The core challenge of SA-OWOD lies in resolving the ambiguity of unlabeled regions.  
KTRM reduces false-negative supervision caused by missing annotations, preserving K-mAP stability under sparsity. 
DDTG strengthens unknown modeling by explicitly regularizing semantically inconsistent proposals. 
The reduced U-Recall degradation and performance gap \AJ{validate} that explicitly modeling both mechanisms is essential for robust detection in sparsely annotated open-world environments.








\subsection{Ablation Studies}


We conduct ablation studies under the Hard setting to analyze the individual contributions of the proposed KTRM and DDTG, as well as the effect of key thresholds in DDTG. Moreover, we provide a qualitative comparison of the results with those of a baseline model.


\begin{table}[t]
\centering

\begin{minipage}[t]{\textwidth}
\centering
\setlength{\tabcolsep}{5pt}
\renewcommand{\arraystretch}{1.05}

\caption{\textbf{Ablation study on KTRM and DDTG under the Hard setting\AJ{.}}}
    \resizebox{0.8\textwidth}{!}{%
    \begin{tabular}{l c c | c c | c c}
    \toprule
    Task 1 & KTRM & DDTG
    & K-mAP & $\Delta$
    & U-Recall & $\Delta$ \\
    \midrule
    
    CROWD
    & \ding{55} & \ding{55}
    & 49.29 & --
    & 49.80 & -- \\
    
    CROWD+KTRM
    & \ding{51} & \ding{55}
    & 51.90 & +2.61
    & 50.66 & +0.86 \\
    
    CROWD+DDTG
    & \ding{55} & \ding{51}
    & 51.08 & +1.79
    & 50.26 & +0.46 \\
    
    Ours
    & \ding{51} & \ding{51}
    & \textbf{54.09} & \textbf{+4.80}
    & \textbf{51.85} & \textbf{+2.05} \\
    
    \bottomrule
    \end{tabular}%
    }
    \label{abl:KTRM_DDTG}

\end{minipage}

\hfill

\begin{minipage}[t]{\textwidth}
\centering
\setlength{\tabcolsep}{6pt}
\renewcommand{\arraystretch}{1.05}

\caption{\textbf{Ablation study on $\tau_o$ and $\tau_s$ in DDTG under the Hard setting\AJ{.}}}
\label{tab:threshold}
    \resizebox{0.7\textwidth}{!}{%
    \begin{tabular}{cccc|cccc}
    \toprule
    \multicolumn{4}{c|}{Objectness threshold ($\tau_o$)}
    &
    \multicolumn{4}{c}{Similarity threshold ($\tau_s$)} \\
    \cmidrule(lr){1-4} \cmidrule(l){5-8}
    
    $\tau_o$ & $\tau_s$ & K-mAP & U-Recall
    &
    $\tau_o$ & $\tau_s$ & K-mAP & U-Recall \\
    \midrule
    
    0.1 & 0.95 & 52.97 & 47.68
    &
    \textbf{0.2} & \textbf{0.95} & \textbf{54.09} & \textbf{51.85} \\
    
    \textbf{0.2} & \textbf{0.95} & \textbf{54.09} & \textbf{51.85}
    &
    0.2 & 0.90 & 52.24 & 46.26 \\
    
    0.3 & 0.95 & 53.62 & 51.40
    &
    0.2 & 0.85 & 52.53 & 50.46 \\
    
    \bottomrule                                             
    \end{tabular}%
    }
    \label{abl:obj_sim}

\end{minipage}
\end{table}


\noindent\textbf{Effect of KTRM and DDTG.}
As shown in \cref{abl:KTRM_DDTG}, we perform an ablation study to analyze the individual contributions of the proposed KTRM and DDTG under the Hard setting. Starting from the CROWD baseline, we progressively enable each module to examine their impact on both known object detection performance (K-mAP) and unknown object recall (U-Recall).

When only KTRM is applied, improvements are mainly observed in known category detection, indicating that KTRM enhances discriminative representation learning and stabilizes training. Applying only DDTG also improves performance, suggesting that the proposed target generation mechanism contributes to boundary refinement and unknown exploration\AJ{.}

When both modules are jointly applied, consistent improvements are observed across both K-mAP and U-Recall. 
This demonstrates that KTRM and DDTG play complementary roles: KTRM improves feature quality for known categories, while DDTG facilitates unknown object exploration. The combined design achieves a better balance between known recognition and unknown discovery.


\begin{figure}[t]
    \centering
    \includegraphics[width=1\linewidth]{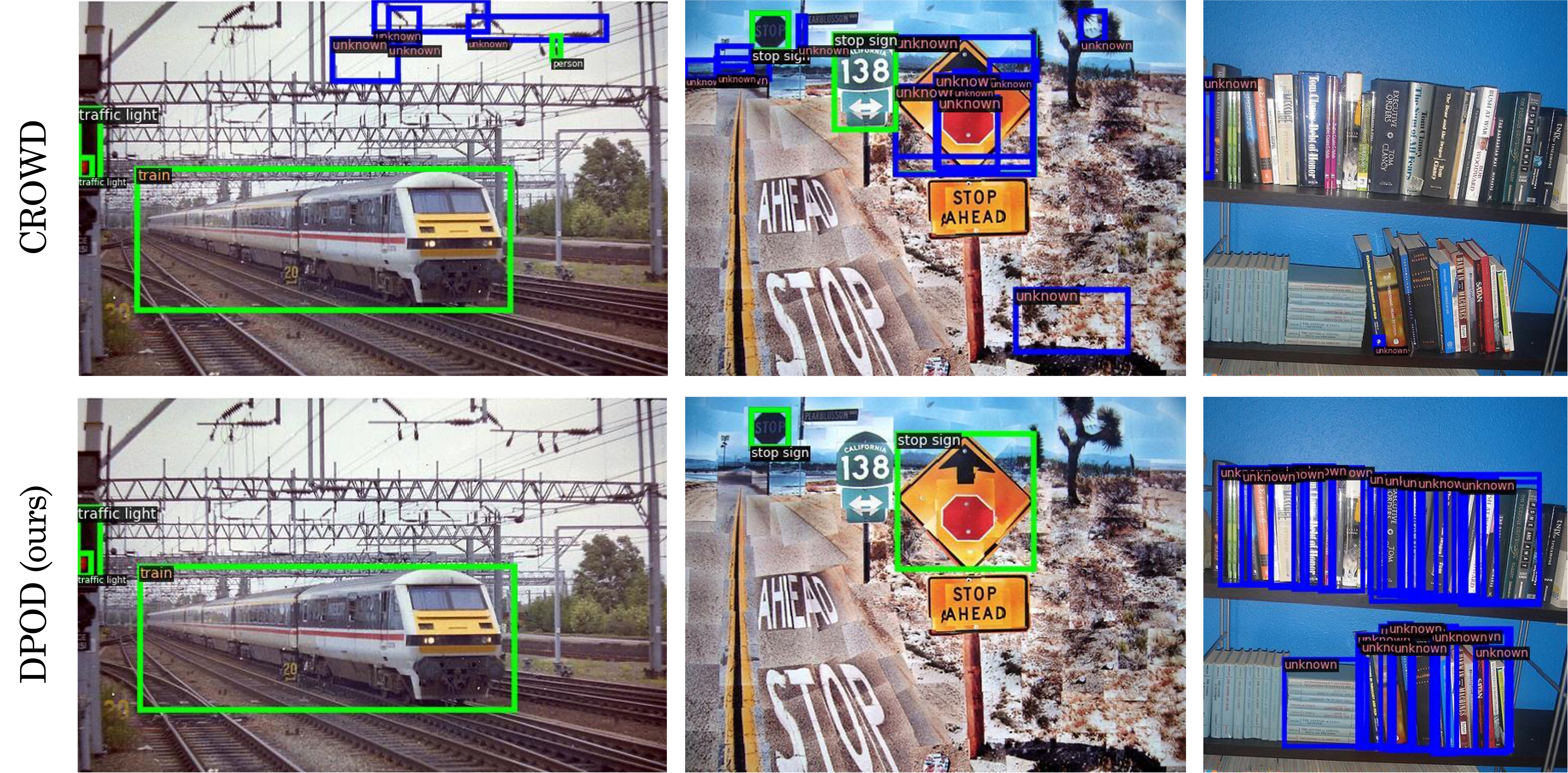}
    \caption{\textbf{Qualitative Comparison on Hard Task 3.} 
    We compare CROWD~\cite{majee2025looking} and our method in terms of known and unknown detections. \textcolor{blue}{Blue} denotes unknown objects, and \textcolor{green}{green} denotes known objects.}
    \label{fig:visualization}
\end{figure}


\noindent\textbf{Sensitivity to Objectness and Similarity Thresholds.}
We further analyze the sensitivity of DDTG to the objectness threshold ($\tau_o$) and similarity threshold ($\tau_s$), as shown in \cref{abl:obj_sim}. 
When fixing $\tau_s=0.95$, increasing $\tau_o$ from 0.1 to 0.2 significantly improves both K-mAP and U-Recall, indicating that moderate filtering of low-objectness proposals helps suppress noise while preserving informative candidates. 
However, further increasing $\tau_o$ to 0.3 slightly degrades performance, suggesting that overly strict filtering may discard useful unknown candidates.

When fixing $\tau_o=0.2$, increasing $\tau_s$ enlarges the set of proposals regarded as semantic disagreement.
Although a larger disagreement set may potentially introduce noisy candidates, the results indicate that the additional proposals provide informative supervisory signals rather than degrading detection quality. 
We conjecture that richer disagreement cues encourage the model to learn a clearer separation between known and unknown representations, which benefits not only unknown recall but also known-class detection performance.
Reducing $\tau_s$ limits the diversity of disagreement candidates, leading to consistent drops in both metrics. These findings suggest that sufficiently diverse semantic disagreement signals are important for stable and effective open-world detection.

\noindent\textbf{Qualitative Results\AJ{.}}
The qualitative results in \cref{fig:visualization} visually support the qualitative improvements reported in Hard Task 3. In CROWD~\cite{majee2025looking}, known objects are sometimes misclassified as unknown, and some unknown instances fail to be clearly activated. In contrast, our method reduces ambiguity around object boundaries and produces more consistent detections. This suggests that our approach achieves improved stability even under incomplete supervision.

\section{Conclusion}


In this work, we introduced \textit{Sparsely Annotated Open-World Object Detection (SA-OWOD)}, where unlabeled known and unknown objects co-exist. In this setting, unlabeled regions are inherently ambiguous and cannot be reliably treated as either unlabeled known or unknown. To address this challenge, we proposed \textit{Dual-Perspective Object Discovery (DPOD)}, a unified framework that explicitly separates unlabeled regions into recoverable known objects and genuinely unknown ones. By explicitly modeling and separating unlabeled known and unknown instances, DPOD effectively resolves this ambiguity. Extensive experiments demonstrate consistent improvements in both known-class detection and unknown recall. Overall, SA-OWOD provides a practical benchmark for detection under ambiguous supervision and lays the foundation for robust open-world detection in real-world scenarios. 

\noindent\HJ{\textbf{Limitations.} Despite these advances, DPOD currently relies on fixed objectness and score thresholds ($\tau_o$, $\tau_s$) for proposal filtering. In open-world scenarios, object characteristics vary significantly across tasks, and certain categories — such as accessories (e.g., ties) — tend to yield lower objectness scores. Enforcing a fixed threshold under such noisy conditions is suboptimal and may introduce progressive learning bottlenecks. Replacing these fixed heuristics with task-adaptive thresholds remains a promising direction for future work.}


\section*{Acknowledgements}
%
This work was supported in part by the National Research Foundation of Korea (NRF) grant funded by the Korea\AJ{n} government (MSIT) (RS-2024-00358935, 90\%) and in part by the Institute of Information \& Communications Technology Planning \& Evaluation (IITP) under the Artificial Intelligence Convergence Innovation Human Resources Development grant funded by the Korea government (MSIT) (IITP-2026-RS-2023-00254177, 10\%).

%
%
\bibliographystyle{splncs04}
\bibliography{main}
\end{document}